%% file: template.tex
\documentclass{Interspeech}

\interspeechcameraready

\usepackage{tipa}
\usepackage[utf8]{inputenc}
\usepackage{graphicx}
\usepackage[T1]{fontenc}
\usepackage{booktabs}
\usepackage{multirow}
\usepackage{color, colortbl}
\usepackage{caption}
\usepackage{subcaption}
\usepackage{booktabs}
\usepackage{bm}
\usepackage{amsmath}
\usepackage{cleveref}
\usepackage{tcolorbox}
\usepackage{amsmath}
\usepackage{amssymb}
\usepackage{xcolor} 
\usepackage{arydshln} 
\usepackage{xspace}
\usepackage{array}
\usepackage{lingmacros}

\usepackage{hyperref}

\newcommand*{\afrihubert}{AfriHuBERT\xspace}
\newcommand*{\mhubert}{mHuBERT-147\xspace}

\newcommand*{\fleurs}{FLEURS\xspace}
\newcommand*{\yoruba}{Yor\`ub\'a\xspace}

\title{\afrihubert: A self-supervised speech representation model for African languages}

\author[affiliation={1}]{Jesujoba O.}{Alabi}
\author[affiliation={2}]{Xuechen}{Liu}
\author[affiliation={1}]{Dietrich}{Klakow}
\author[affiliation={2}]{Junichi}{Yamagishi}

\affiliation{Saarland University}{Saarland Informatics Campus}{Germany}
\affiliation{}{National Institute of Informatics}{Japan}
\email{\{jalabi,dietrich.klakow\}@lsv.uni-saarland.de, \{xuecliu,jyamagis\}@nii.ac.jp}


\usepackage{comment}

\begin{document}

\maketitle

\begin{abstract}
    
    In this work, we present \afrihubert, an extension of \mhubert, a compact self-supervised learning (SSL) model pretrained on 147 languages. While \mhubert covered 16 African languages, we expand this to 1,226 through continued pretraining on 10K+ hours of speech data from diverse sources, benefiting an African population of over 600M. We evaluate \afrihubert on two key speech tasks, Spoken Language Identification (SLID) and Automatic Speech Recognition (ASR), using the FLEURS benchmark. Our results show a +3.6\% F1 score improvement for SLID and a -2.1\% average Word Error Rate (WER) reduction for ASR over \mhubert, and demonstrates competitiveness with larger SSL models such as MMS and XEUS. Further analysis shows that ASR models trained on \afrihubert exhibit improved cross-corpus generalization and are competitive in extremely low-resource ASR scenarios.
\end{abstract}

\section{Introduction}

\label{sec:intro}
Self-supervised learning (SSL)-based speech representation  models such as HuBERT~\cite{hubert2021hsu}, XLS-R~\cite{babu2021xlsr}, and WavLabLM~\cite{chen2023joint} have become an important component in the development of various speech-related applications, such as automatic speech recognition (ASR)~\cite{Pascual2019,chung2021w2v,zhang2023google}, speech synthesis~\cite{gong2023zmm}, speech translation~\cite{seamless2023}, and spoken language understanding (SLU)~\cite{peng2023study}. These models, trained on vast amounts of unlabeled data, are designed to capture the nuances of different languages, enabling robust and accurate performance across diverse tasks.  

Existing SSL models can be categorized as either monolingual or multilingual. Most monolingual SSL models are trained exclusively on English~\cite{hubert2021hsu,baevski2020wav2vec}
, with only a few multilingual SSL models available, such as mHuBERT~\cite{lee2021textless}, w2v-XLSR~\cite{babu2021xlsr}, 
\mhubert~\cite{boito2024mhubert}, and MMS~\cite{pratap2024scaling}, which cover up to a thousand languages. While these models include some African languages, English and a few other high-resource languages often dominate the training data due to the abundance of available resources. Despite Africa's rich linguistic diversity, African languages remain relatively underrepresented. This lack of representation creates significant challenges in the building of robust speech-based dialog systems on the African continent, where thousands of languages and dialects coexist.

Some of the recent multilingual SSL models, such as MMS~\cite{pratap2024scaling}, w2v-BERT 2.0~\cite{seamless2023}, and XEUS~\cite{chen-etal-2024-towards-robust}, have demonstrated strong performance across different languages and tasks. However, these models tend to be large, with over 300 million parameters--- making them computationally expensive and challenging to deploy in resource-constrained environments.

In contrast, \mhubert, which is trained on 147 languages, including 16 African languages, offers a compact yet competitive alternative.
Although it is competitive on benchmarks like ML-SUPERB~\cite{Shi2023mlsuperb}, it lags behind on the \fleurs~\cite{conneau2023fleurs} benchmark for most languages, including most African languages. To address this gap, we introduce \afrihubert, the first massively multilingual and compact African-centric SSL model, built by extending \mhubert (95M parameters) through continued pretraining. \afrihubert is trained on a diverse dataset of 1,226 African languages and dialects, plus four widely spoken languages in Africa: Arabic, English, French, and Portuguese. The pretraining dataset is sourced from diverse domains, ensuring comprehensive phonetic and linguistic representation of the African context.


We evaluate our model on two downstream tasks, spoken language identification (SLID) and ASR, using \fleurs. \afrihubert significantly outperforms existing small SSL models on both tasks, including for languages with adaptation data solely from religious domains. This narrows the performance gap between \mhubert and large SSL models for African languages and, thereby highlighting the importance of tailored pretraining for speech representation in African languages. Our contributions are as follows.
\begin{enumerate}
    \item We aggregate more than 10,000 hours of speech, covering more than 1,200 African languages and dialects.
    \item We introduce \afrihubert, a multilingual SSL model for African languages, comparing continued pretraining and training from scratch over one iteration.
    \item We evaluate AfriHuBERT against other multilingual SSL models on SLID and ASR, analyzing its predictions for both tasks.
\end{enumerate}
We have released the pre-trained models of \afrihubert,\footnote{\url{https://huggingface.co/ajesujoba/AfriHuBERT}}\footnote{\url{https://doi.org/10.5281/zenodo.15531766}} along with the codebase.\footnote{\url{https://github.com/nii-yamagishilab/AfriHuBERT}}



\section{Data and Pre-processing}

\input{tables/data_source}
\begin{figure*}[t]
  \centering
  \includegraphics[width=\linewidth, height=0.16\textheight, keepaspectratio]{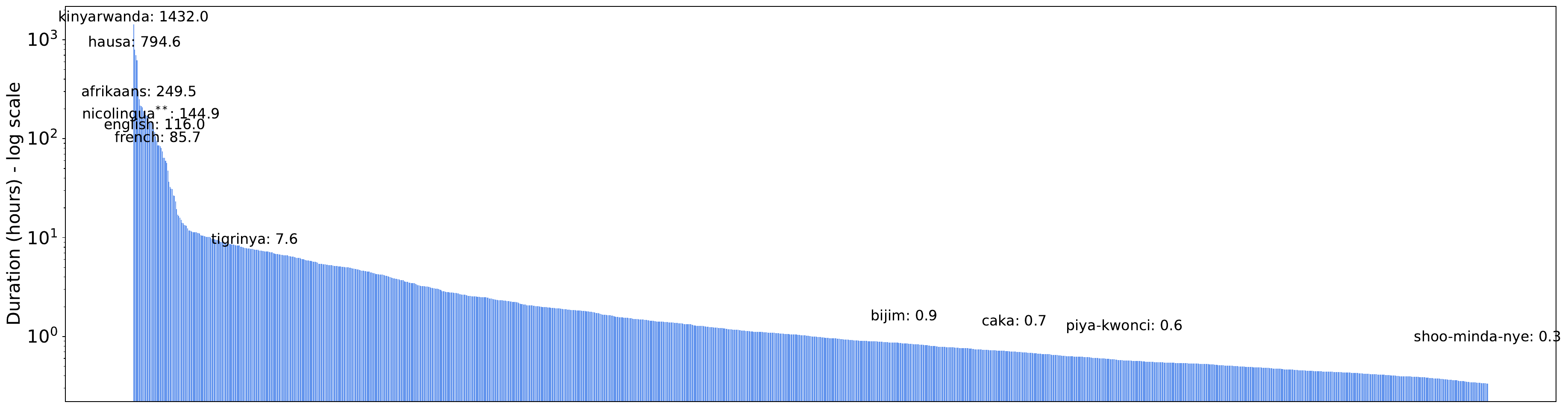}
  \vspace{-10pt}
  \caption{Language-wise duration distribution of the aggregated data after preprocessing. Nicolingua** is not a language, but a collection of speech data for 11 languages, including ten from Guinea (a country in West Africa).} 
  \label{fig:lang_dist} \vspace{-12pt} 
\end{figure*}
\subsection{Continued pretraining dataset}
We aggregate data from various speech datasets across 1,230 languages, including 1,226 African languages and Arabic, English, French, and Portuguese. The four non-African languages are included to help preserve the model’s ability on these languages and on their African-accented varieties. The data is gathered from 11 major sources: BibleTTS~\cite{meyer2022bibletts}, Congolese Speech Radio Corpus (CSRC)~\cite{kimanuka2024speech}, Jesus Dramas~\cite{chen-etal-2024-towards-robust}, Kallaama~\cite{kallaama2024dataset}, Mozilla Common Voice (MCV)
~\cite{ardila-etal-2020-common},\footnote{version 17.0 hosted on Huggingface} MMS ulab v2~\cite{pratap2024scaling,chen-etal-2024-towards-robust}, NCHLT~\cite{barnard2014nchlt,badenhorst2022nchlt}, Nicolingua radio corpus~\cite{doumbouya2021usingradio},  NaijaVoices~\cite{emezue2025naijavoicesdatasetcultivatinglargescale}
, VoxLingua107~\cite{valk2021voxlingua107}, and Zambezi voice~\cite{sikasote23_interspeech}. We filter Jesus Dramas and MMS ULAB v2 for African languages using GPT-4o\footnote{Version dated 2024-08-06.} and Glottolog,\footnote{\url{https://github.com/glottolog/glottolog}} respectively; the former uses language names, the latter ISO 639-3 codes. Combining these data sources yields speech samples for 1,439 languages, including the four non-African languages. \Cref{tab:data_source} summarizes the data sources and their properties.

\subsection{Evaluation dataset}
For downstream evaluation, we use the Sub-Saharan Africa (SSA) subset of the \fleurs dataset, which includes 20 languages from our pretraining data. We also include Kinyarwanda \fleurs,\footnote{\url{https://huggingface.co/datasets/mbazaNLP/fleurs-kinyarwanda}} and Arabic, English, French, and Portuguese, totaling 25 languages. We focus on SLID and multilingual ASR as downstream tasks.

\subsection{Data preprocessing}
All data including \fleurs were converted to single-channel audio and downsampled to 16 kHz. CSRC, Jesus Dramas, and MMS ulab v2 were segmented using WebRTC VAD.\footnote{\url{https://github.com/wiseman/py-webrtcvad}} The Kallaama dataset was split into manageable segments using transcription files provided by the authors.
The noise in VoxLingua107 was filtered using the manifest from~\cite{boito2024mhubert}. 
Additionally, we removed audio segments shorter than 1 second or longer than 30 seconds and excluded languages with less than 20 minutes of audio. As a result, the pretraining dataset contains over 10,000 hours of audio, covering 1,230 out of the 1,439 languages originally gathered. \Cref{fig:lang_dist} shows a skewed duration distribution, with Kinyarwanda accounting for over 10\% of the total, while many languages have less than 10 hours of speech. 

\section{AfriHuBERT: Setup and Training}
\label{sec:afrihubert}
We train \afrihubert by extending \mhubert with the aggregated data, using multilingual adaptive finetuning (MAFT)~\cite{tang2020multilingual,alabi-etal-2022-adapting}, a process of continued pretraining on multiple languages at once.
Given the strong capabilities of \mhubert, we use a one-iteration adaptation strategy~\cite{xu2024seamless}. Our objective is to answer two questions (1) \emph{Can massively pre-trained \mhubert effectively generalize to African languages?} (2) \emph{How effective is training \afrihubert from scratch using quality discrete targets from the pre-trained \mhubert without refinement?} Hence, we train three versions of \afrihubert. The first two require MAFT on \mhubert using its original discrete targets from the k-means model, while the other trains the k-means model on African language datasets to obtain AfriHuBERT-\emph{o} and AfriHuBERT-\emph{n}, respectively. Lastly, we train AfriHuBERT-\emph{s} from scratch for one iteration using the new discrete targets. The new k-means model is trained using representations from the 9th layer of \mhubert with Faiss-based clustering~\cite{douze2024faisslibrary}. We sampled up to 1 hour of speech data from each language and merged all samples to train the clustering model.\footnote{Note that loading the entire dataset into memory is infeasible.}

To address language imbalance and ensure the model learns from underrepresented languages and dialects, we upsampled the aggregated data using temperature sampling with a multinomial distribution: 
\begin{equation}
    \vspace{-0.5pt} 
    q_i = \frac{p_i^\alpha}{\sum_{j=1}^{D} p_j^\alpha}, \quad \text{where} \quad 
    p_i = \frac{d_i}{\sum_{j=1}^{D} d_j}.
    \vspace{-0.5pt} 
\end{equation}
Here $d$ is each language duration, $D$ is the total number of languages, $p$ is the probability of the language, and $\alpha$ is a temperature parameter that we set to 0.8.\footnote{It is computationally expensive to test all possible values of $\alpha$.} We exclude English, Arabic, French, Portuguese, and audio data in Nicolingua.\footnote{These exclusions were due to the inability to separate the audio data into the ten respective languages including French.} Before upsampling, we allocate 10 minutes of audio for languages with less than 2 hours of data and 30 minutes for others as the validation set. We also include the original Nicolingua validation split. 
%
We train the models for 100K steps on upsampled data using the original HuBERT implementation within Fairseq~\cite{ott2019fairseq}. Training uses a maximum of 128K tokens per batch, an update frequency of 64, and is optimized with a learning rate of 5e$^{-5}$ and 32K warm-up steps.\footnote{lr = 5e$^{-3}$ when training from scratch.} The final models are selected based on the checkpoint with the lowest validation loss. Each model is trained using 4 NVIDIA A100 40GB GPUs.

\section{Supervised Finetuning Setup}
We evaluate \afrihubert models on SLID and multilingual ASR via supervised finetuning with FLEURS. Alongside \mhubert, we evaluate a few other SSL models, including SSA-HuBERT~\cite{caubriere2024africacentric} (95M params, trained on 20 African languages/dialects). We also fine-tune larger SSL models: XLSR-128~\cite{babu2021xlsr} and MMS~\cite{pratap2024scaling} (316M), as well as w2v-BERT 2.0~\cite{seamless2023} and XEUS~\cite{chen-etal-2024-towards-robust} (580M). The latter two are not directly comparable due to their size and extensive pretraining; XEUS includes \fleurs in its training data, while w2v-BERT 2.0's pretraining data is undisclosed.

For SLID, we fine-tune on 25 \fleurs languages using an attentive static pooling layer, followed by a 512-D and 25-D softmax layer. Models are trained for 20 epochs with 3 random seeds; we report average F1 across all languages, both including and excluding the 4 non-African ones. To address imbalance, we cap each language at 1,030 samples, matching Afrikaans.

For multilingual ASR, models are fine-tuned jointly on all languages using CTC loss, without an external language model. The full fine-tuning setup uses a 3-layer FFN (1024 neurons, and LeakyReLU activation). We apply a 432-size character vocab from SentencePiece~\cite{kudo-richardson-2018-sentencepiece}. Training runs for 30 epochs with 3 random seeds, and we report the average overall WER. To ensure balanced training, we sample three hours of audio per language and merge the data across all 25 languages.

Following~\cite{boito2024mhubert}, we optimize fine-tuning for both tasks using Adam for the speech encoder, selecting the best learning rate from ${1e^{-3}, 1e^{-4}, 1e^{-5}}$. For SLID, the FFN uses Adam with a fixed learning rate of 0.001; for ASR, it uses Adadelta with a learning rate of 1.0, as implemented within SpeechBrain~\cite{speechbrain}.\footnote{We adapted the IEMOCAP and VoxLingua107 SpeechBrain recipes for SLID, and the DVoice recipe for ASR.} All models are trained on a single NVIDIA A100 GPU (40GB/80GB) with batch sizes of 32 (SLID) and 16 (ASR), using gradient accumulation as needed.
 

\section{Results}

\Cref{tab:main_result} shows the SLID and ASR results, including average F1 and WER across all 25 languages and specifically for African languages. The following paragraphs summarize our key findings.

\input{tables/main_result}

\textbf{\mhubert is a strong, compact, multilingual SSL baseline}. Overall, \mhubert outperforms SSA-HuBERT—a multilingual model of the same size—with a lower average WER, while SSA-HuBERT performs better in SLID. At the language level, we observed that SSA-HuBERT achieves better F1/WER on Hausa and Swahili, perhaps benefiting from pretraining on large datasets for both languages.

\textbf{Performing MAFT on \mhubert led to improved performance on African languages.} On average, both AfriHuBERT-\emph{o} and AfriHuBERT-\emph{n} outperform the other two small SSL models on both tasks, while achieving comparable performance. Compared to \mhubert, both models perform better on all African languages except Arabic, English, French, and Portuguese, which were dominant during pretraining but underrepresented during adaptation. Languages like Luo and Kimbundu, which were not present during pretraining and introduced only during adaptation with a few hours of religious data, show improvements over \mhubert.
Also, AfriHuBERT-\emph{s} outperforms all small models on SLID but is not competitive to \mhubert or other \afrihubert{s} for ASR. We hypothesize that training AfriHuBERT-\emph{s} for longer steps might improve its representation to the point that it can match \mhubert's performance on African languages. 

\textbf{w2v-BERT 2.0 is a large competitive model}. Among large SSL models, w2v-BERT 2.0, trained on more than 4.5M hours of audio, achieves the best overall performance on ASR due to its size and data volume, while XEUS delivers the best SLID performance but significantly lags behind w2v-BERT 2.0 in ASR. MMS and w2v-XLSR, with similar parameter counts and pretraining data, perform competitively, with MMS showing a slight improvement.

\section{Analysis \& Discussion }
Going forward, we focus on AfriHuBERT-\emph{n}, now called \afrihubert. We analyze its failure cases on both tasks, assess cross-corpus ASR generalization, and evaluate its performance in extremely low-resource and multi-dialect ASR, comparing it to \mhubert and MMS.

\textbf{SLID confusion matrix for \afrihubert}:
Inspecting \afrihubert's confusion matrix reveals that geographically close languages are often misclassified as each other. For example, 40\% of the audio samples speaking Zulu, a South African language, are misclassified as Xhosa on average. However, this miss-classification does not occur in the reverse direction. We hypothesize that this stems from training data artifacts or linguistic similarities, which future work can explore. Similarly, Fulfude, spoken in West and Central Africa, is misclassified as Hausa, Somali, or Wolof, which are languages from overlapping regions.


\textbf{Error analysis of the multilingual ASR output:} 
Based on the ASR results, we analyze system outputs, with a focus on \afrihubert. Using \yoruba (the language with the second-highest WER) as a case study, we identify data quality issues, likely due to error propagation from the FLORES-101~\cite{goyal-etal-2022-flores} dataset, the source of \fleurs. A manual inspection of the \yoruba transcriptions shows that they did not follow the standard \yoruba orthography with instances without diacritics, or a mixture of diacritics and no diacritics. For example: 
\enumsentence{\label{ex:first-example} \small\textbf{Groundtruth transcription:}  \textcolor{orange}{won se ikede naa leyin ti \underline{trumpi} ba \underline{aare toki resep tayipi edogani} lori ago} \\
\textbf{When diacritized:} \textcolor{blue}{w{\d {\' o}}n {\d s}e {\` i}k{\' e}de n{\' a}{\` a} l{\d {\' e}}y{\` i}n t{\' i} \underline{trumpi} b{\' a} \underline{{\` a}{\` a}r{\d e} toki resep tayipi edogani} l{\' o}r{\' i} ago} \\
\textbf{Translation:} \textcolor{purple}{they made the announcement after trump had president toki resep tayipi edogani on a phone call}  \\
\textbf{\afrihubert:} w{\d {\' o}}n se {\` i}k{\' e}de n{\' a}{\` a} l{\d e}y{\` i}n t{\' i} tromp b ar{\d e} toki recept tayip{\` a} {\` e}d{\` o}g{\' a}ni l{\' o}r{\' i} ago 
}

Example \ref{ex:first-example} shows a ground-truth transcription that does not follow \yoruba orthography and is completely undiacritized, while \afrihubert's output is partially diacritized. These inconsistencies, especially in the \fleurs training data, likely contribute to the \yoruba ASR models' high WER. Beyond diacritics, the models also struggle with transcribing named entities. Future work should further audit \fleurs transcriptions and correct these errors, similar to \fleurs-R~\cite{ma24c_interspeech}, which focused on improving \fleurs's audio quality.

\input{tables/result_asr_low}

\textbf{Cross-corpus ASR generalization of \afrihubert:} Next, we evaluate the ASR models from \afrihubert, \mhubert, and MMS—three multilingual HuBERT-style models—on another ASR corpus to assess their cross-corpus generalization. For this, we used the MCV~\cite{ardila-etal-2020-common} test split,\footnote{version 17.0} which covers eight of the 21 African languages they were originally trained on. The CER and WER results presented in \Cref{tab:analysis2} show that \afrihubert, on average, generalizes better out-of-domain and outperforms \mhubert and MMS with WERs of 68.8\% and 65.2\%, respectively. Specifically, \afrihubert achieves a WER of less than 60\% in three languages (Afrikaans, Hausa, Swahili) and of less than 70\% for Igbo, Kinyarwanda and Luganda. In contrast, Amharic (using a non-Latin script) and \yoruba (using Latin script with diacritics) have WERs of 81.0\% and 81.2\%, respectively. 

\textbf{Evaluating \afrihubert on low-resource multilingual ASR:}
Furthermore, we evaluate the three SSL models in extremely low-resource settings. Using the experimental setup from \Cref{sec:afrihubert}, we fine-tune the models for multilingual ASR with 10 and 30 minutes of audio data per language, respectively, and evaluate their performance on African languages only.

\begin{figure}[t]
  \centering
  \includegraphics[width=\linewidth, height=0.28\textheight, keepaspectratio]{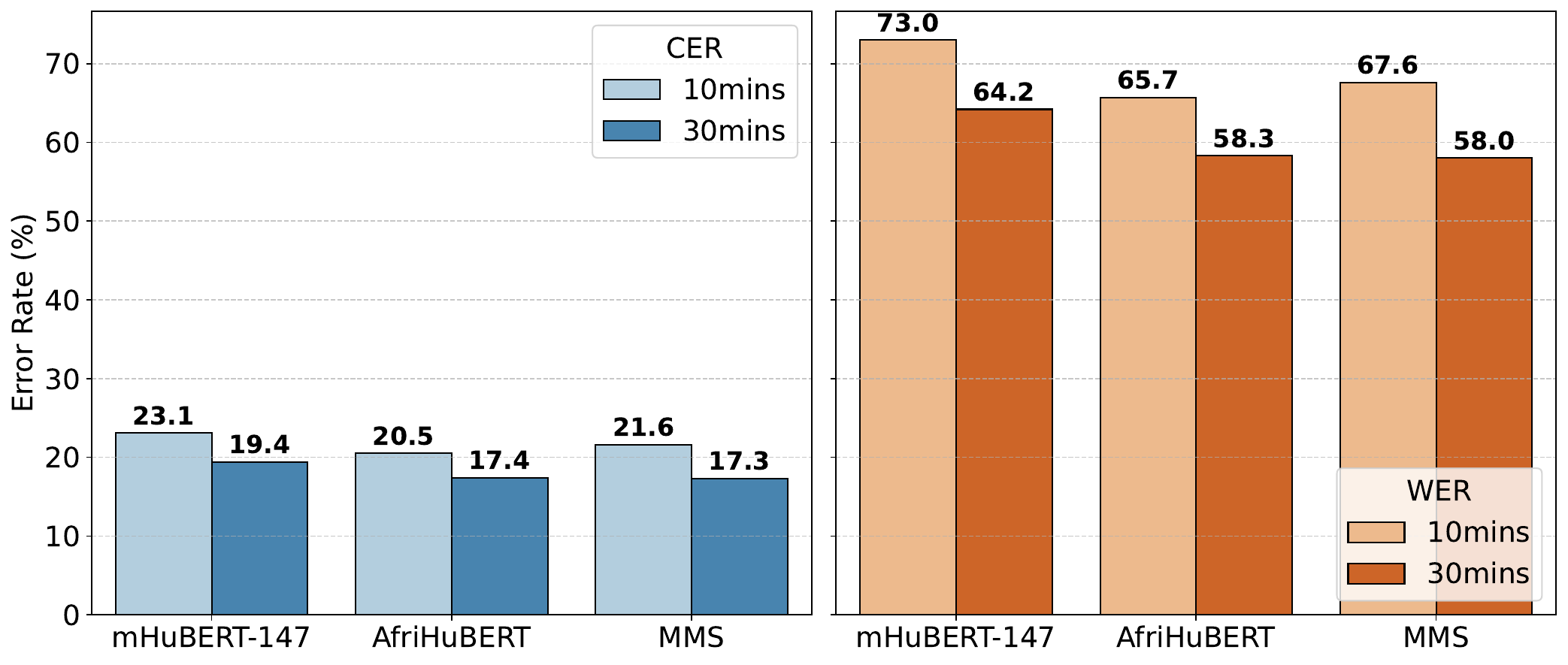}
  \vspace{-20pt}
  \caption{ASR performance in extremely low-resource ASR scenarios.}
  \label{fig:few_shot} \vspace{-14pt} 
\end{figure}

The results in \Cref{fig:few_shot} show that, on average, \afrihubert outperforms \mhubert in both settings and remains competitive with MMS. In the 10-minute setting, \afrihubert achieved a WER of 65.7\% and a CER of 20.5\%, compared to \mhubert’s 73.0\% and 23.1\%, respectively. In the 30-minute setting, WERs considerably decreased across all three models. However, for most languages, WERs remained above 60\%, with only a few exceptions, highlighting the need for more data to improve ASR performance for these languages.

\input{tables/result_yo}

\textbf{Multi-dialect \yoruba ASR evaluation:} Lastly, given the flaws identified in \yoruba FLEURS, we address whether our findings, particularly the observed improvements and competitiveness by \afrihubert, can be trusted. We train a multi-dialect \yoruba ASR model on the well-curated {YOR{\` U}LECT}~\cite{ahia-etal-2024-voices} dataset using a similar setup as before but with a sentencepiece character vocabulary of size 63. We focus on three dialects: Standard \yoruba, Ife, and Ilaje. Our results in \Cref{tab:asr_yoruba} show that, on average and across dialects, \afrihubert outperforms \mhubert, and is competitive to MMS. We hypothesize that models perform best on the standard dialect due to its abundant resources, while Ife is the most challenging due to its scarcity. These results confirm that, despite \afrihubert's compact size, it is still competitive.

\section{Conclusion}
In this work, we created \afrihubert by extending \mhubert to 1,226 African languages via MAFT on speech data aggregated from various sources, including these languages and four widely spoken non-indigenous languages in Africa. We evaluated both compact SSL models such as \mhubert and \afrihubert together with some other large multilingual SSL models on both SLID and ASR tasks and found that \mhubert is a strong multilingual SSL baseline. However, \afrihubert which is an extension of \mhubert, outperforms other SSL models on average for both tasks. 
In future work, we plan to upscale \afrihubert, enhancing its generalizability to better accommodate unseen African languages and dialects.

\section{Acknowledgements}
This work was conducted during the first author’s internship at NII, Japan.  This study is partially supported by JST AIP Acceleration Research (JPMJCR24U3).  Part of this study was carried out using the TSUBAME4.0 supercomputer at the Institute of Science Tokyo. Also, we thank Xin Wang, Badr M. Abdullah, Siyang Wang, Wanying Ge, David Adelani, and Aravind Krishnan for their helpful feedback. 


\bibliographystyle{IEEEtran}
\bibliography{mybib}

\end{document}

%% file: tables/data_source.tex
\begin{table*}[th]
\caption{Datasets used for training \afrihubert (after filtering and preprocessing the aggregated data), the amount of languages covered, the total duration, the domain of the data, the speech type, and licenses.} 
\vspace{-8pt} 
\label{tab:data_source}
  \centering
  \scalebox{0.90}{
  \begin{tabular}{p{4cm}|rrccc}
    \toprule
    \textbf{Name} & \#\textbf{Languages} & \textbf{Duration (h)}  & \textbf{Domain} & \textbf{Type} & \textbf{License} \\ 
    \hline
    BibleTTS ~\cite{meyer2022bibletts} & 6 & 357.6 & Religious & Read & CC BY-SA 4.0 \\ 
    CSRC~\cite{kimanuka2024speech} & 3 & 0.1 & General & Radio & CC-BY \\
    Jesus Dramas ~\cite{chen-etal-2024-towards-robust} & 88 & 99.6 & Religious & Read & CC BY-NC-SA 4.0 \\
    Kallaama~\cite{kallaama2024dataset} & 3 & 124.9 & Agriculture & Spontaneous & CC BY-SA 4.0 \\
    MCV~\cite{ardila-etal-2020-common} & 4 & 1606.1 & General & Read & CC-0  \\ 
    MMS ulab v2~\cite{pratap2024scaling,chen-etal-2024-towards-robust} & 1230 & 2835.4 & Religious & Read & CC BY-NC-SA 4.0 \\
    NaijaVoices~\cite{emezue2025naijavoicesdatasetcultivatinglargescale} & 3 & 1873.9 & General & Read & CC BY-NC-SA 4.0 \\ 
    NCHLT~\cite{barnard2014nchlt,badenhorst2022nchlt} & 10 & 1889.4 & General & Read & CC BY 3.0 \\
    Nicolingua~\cite{doumbouya2021usingradio} & 10 & 142.4 & News & Radio & CC BY-SA 4.0 \\ 
    VoxLingua107~\cite{valk2021voxlingua107} & 13 & 886.4 & General & Spontaneous & CC BY 4.0 \\ 
    Zambezi Voice~\cite{sikasote23_interspeech} & 5 & 176.0 & General & Radio & CC BY-NC-ND 4.0 \\
    \bottomrule
  \end{tabular}
  }
  \vspace{-5pt} 
  
\end{table*}

%% file: tables/main_result.tex
\begin{table}[ht]
  \caption{Performance of SSL models on FLEURS. We report the average F1 (\%) and WER (\%) scores for all languages (avg$_{*}$), and only the 21 African languages (avg). Size refers to each model's parameters (millions), and Dur denotes their pretraining data size (million hours). }
  \label{tab:main_result} \vspace{-8pt} 
\centering
\scalebox{0.88}{
\begin{tabular}{lcc|ll|ll}
\toprule
\textbf{Models} & \textbf{Size}  & \textbf{Dur} & \multicolumn{2}{c}{\textbf{SLID}(F1)$\uparrow$}  & \multicolumn{2}{c}{\textbf{ASR}(WER)$\downarrow$}  \\ 
 {} & {(M)}  & {M(h)} & avg$_{*}$ & avg  & avg$_{*}$ & avg      \\
 \midrule
  \multicolumn{7}{l}{\textbf{Small SSL}} \\
\mhubert  & 95 & $9e^{-2}$  & 88.0   & 85.8  & 50.4  & 52.1     \\ 
SSA-HuBERT & 95 & $6e^{-2}$    & 89.6   & 88.0   & 56.6   & 56.2   \\ 
\rowcolor{blue!20}
AfriHuBERT-\emph{s}  & 95 & $1e^{-2}$   & 93.2  & 92.0  & 54.2  & 52.9   \\ 
\rowcolor{blue!20}
AfriHuBERT-\emph{o} & 95 & $1e^{-2}$    & 90.3  & 88.9  & 48.4  & 49.3   \\ 
\rowcolor{blue!20}
AfriHuBERT-\emph{n} & 95 & $1e^{-2}$    & 91.6  & 90.0  & 47.9  & 48.7   \\ 
\midrule
\multicolumn{7}{l}{\textbf{Large SSL}} \\
w2v-XLSR   & 317 & $4.4e^{-1}$  & 80.3  & 78.2  & 46.2 & 49.4 \\ 
MMS   & 317 & $4.9e^{-1}$  & 86.3  & 85.6 & 45.6 & 48.0   \\ 
\rowcolor{gray!20}
XEUS   & 577 & $1.1e^{+1}$ & \textbf{96.2} & \textbf{95.5} & 46.5 & 49.5   \\
\rowcolor{gray!20}
w2v-BERT 2.0   & 580 & $4.5e^{+1}$  & 92.7 & 91.3 & \textbf{35.5} &	\textbf{39.3}   \\ 
\bottomrule
\end{tabular}
}
 \vspace{-8pt} 
\end{table}

%% file: tables/result_asr_low.tex
\begin{table}[ht]
  \caption{Cross-corpus generalization of ASR models on MCV.}
  \label{tab:analysis2} \vspace{-10pt} 
\centering
\scalebox{0.70}{
\begin{tabular}{lrrrrrrrr|r}\toprule
&\textbf{afr} &\textbf{amh} &\textbf{hau} &\textbf{ibo} &\textbf{kin} &\textbf{lug} &\textbf{swh} &\textbf{yor} &\cellcolor[HTML]{A8A8A8}Avg \\
\midrule
\multicolumn{10}{l}{\textbf{CER} (\%)}\\
\mhubert & 15.2 &46.2 &17.4 &21.1 &24.6 &18.1 &19.6 &38.8 &\cellcolor[HTML]{A8A8A8}25.1 \\
\afrihubert & 13.2 & \textbf{42.5} & \textbf{14.1} &18.3 & \textbf{22.3} & \textbf{16.6} &17.6 & \textbf{35.8} &\cellcolor[HTML]{A8A8A8} \textbf{22.6} \\
MMS & \textbf{13.1} &48.7 &16.3 & \textbf{17.0} &24.4 &17.2 & \textbf{17.3} &37.2 &\cellcolor[HTML]{A8A8A8}23.9 \\

\hline
\multicolumn{10}{l}{\textbf{WER} (\%)}\\
\mhubert & 53.1 &85.8 &59.4 &62.3 &72.4 &71.3 &59.0 &86.9 &\cellcolor[HTML]{b7b7b7}68.8 \\
\afrihubert & 48.0 & \textbf{81.0} & \textbf{51.1} &60.5 &\textbf{66.5} & \textbf{67.4} & \textbf{52.6} & \textbf{81.2} &\cellcolor[HTML]{b7b7b7} \textbf{63.6} \\
MMS & \textbf{43.6} &83.7 &57.9 & \textbf{56.2} &72.1 &70.6 &53.0 &84.4 &\cellcolor[HTML]{b7b7b7}65.2 \\

\bottomrule
\end{tabular}
}

\vspace{-5pt} 
\end{table}

%% file: tables/result_yo.tex
\begin{table}[ht]
  \caption{Multi-dialect ASR performance comparison on YOR{\` U}LECT (comparing three \yoruba dialects).}
  \label{tab:asr_yoruba} \vspace{-10pt} 
\centering
\scalebox{0.70}{
\begin{tabular}{lrrrrrr|rr}
\toprule
Models &\multicolumn{2}{c}{\textbf{Standard}} &\multicolumn{2}{c}{\textbf{Ife}} &\multicolumn{2}{c}{\textbf{Ilaje}} &\multicolumn{2}{c}{\textbf{Avg}} \\
\midrule
& CER & WER & CER & WER & CER & WER & CER & WER \\\midrule
\mhubert & 11.9 &40.8 &22.4 &65.1 &17.1 &51.0 &17.1 &52.3 \\
\afrihubert & \textbf{11.2} &\textbf{37.7} &\textbf{21.4} &62.9 &16.4 &48.8 &\textbf{16.3} &49.8 \\
MMS &11.4 &38.2 &21.6 & \textbf{62.5} & \textbf{15.8} & \textbf{47.5} & \textbf{16.3} &\textbf{49.4} \\
\bottomrule
\end{tabular}
}
\vspace{-5pt} 
\end{table}

%% file: template.bbl
\begin{thebibliography}{10}
\providecommand{\url}[1]{#1}
\csname url@samestyle\endcsname
\providecommand{\newblock}{\relax}
\providecommand{\bibinfo}[2]{#2}
\providecommand{\BIBentrySTDinterwordspacing}{\spaceskip=0pt\relax}
\providecommand{\BIBentryALTinterwordstretchfactor}{4}
\providecommand{\BIBentryALTinterwordspacing}{\spaceskip=\fontdimen2\font plus
\BIBentryALTinterwordstretchfactor\fontdimen3\font minus \fontdimen4\font\relax}
\providecommand{\BIBforeignlanguage}[2]{{%
\expandafter\ifx\csname l@#1\endcsname\relax
\typeout{** WARNING: IEEEtran.bst: No hyphenation pattern has been}%
\typeout{** loaded for the language `#1'. Using the pattern for}%
\typeout{** the default language instead.}%
\else
\language=\csname l@#1\endcsname
\fi
#2}}
\providecommand{\BIBdecl}{\relax}
\BIBdecl

\bibitem{hubert2021hsu}
W.-N. Hsu, B.~Bolte \emph{et~al.}, ``Hubert: Self-supervised speech representation learning by masked prediction of hidden units,'' \emph{IEEE/ACM transactions on audio, speech, and language processing}, vol.~29, pp. 3451--3460, 2021.

\bibitem{babu2021xlsr}
A.~Babu, C.~Wang \emph{et~al.}, ``Xls-r: Self-supervised cross-lingual speech representation learning at scale,'' \emph{arXiv}, vol. abs/2111.09296, 2021.

\bibitem{chen2023joint}
W.~Chen, J.~Shi \emph{et~al.}, ``Joint prediction and denoising for large-scale multilingual self-supervised learning,'' in \emph{2023 IEEE Automatic Speech Recognition and Understanding Workshop (ASRU)}.\hskip 1em plus 0.5em minus 0.4em\relax IEEE, 2023, pp. 1--8.

\bibitem{Pascual2019}
S.~Pascual, M.~Ravanelli \emph{et~al.}, ``{Learning Problem-Agnostic Speech Representations from Multiple Self-Supervised Tasks},'' in \emph{Proc.INTERSPEECH}, 2019, pp. 161--165.

\bibitem{chung2021w2v}
Y.-A. Chung, Y.~Zhang \emph{et~al.}, ``W2v-bert: Combining contrastive learning and masked language modeling for self-supervised speech pre-training,'' in \emph{2021 IEEE Automatic Speech Recognition and Understanding Workshop (ASRU)}.\hskip 1em plus 0.5em minus 0.4em\relax IEEE, 2021, pp. 244--250.

\bibitem{zhang2023google}
Y.~Zhang, W.~Han \emph{et~al.}, ``Google usm: Scaling automatic speech recognition beyond 100 languages,'' \emph{arXiv preprint arXiv:2303.01037}, 2023.

\bibitem{gong2023zmm}
C.~Gong, X.~Wang \emph{et~al.}, ``Zmm-tts: Zero-shot multilingual and multispeaker speech synthesis conditioned on self-supervised discrete speech representations,'' \emph{IEEE/ACM Trans. Audio, Speech and Lang. Proc.}, vol.~32, p. 4036–4051, Sep. 2024.

\bibitem{seamless2023}
\BIBentryALTinterwordspacing
S.~Communication, L.~Barrault \emph{et~al.}, ``Seamless: Multilingual expressive and streaming speech translation,'' 2023. [Online]. Available: \url{https://arxiv.org/abs/2312.05187}
\BIBentrySTDinterwordspacing

\bibitem{peng2023study}
Y.~Peng, S.~Arora \emph{et~al.}, ``A study on the integration of pre-trained ssl, asr, lm and slu models for spoken language understanding,'' in \emph{2022 IEEE Spoken Language Technology Workshop (SLT)}.\hskip 1em plus 0.5em minus 0.4em\relax IEEE, 2023, pp. 406--413.

\bibitem{baevski2020wav2vec}
A.~Baevski, Y.~Zhou \emph{et~al.}, ``wav2vec 2.0: A framework for self-supervised learning of speech representations,'' \emph{Advances in neural information processing systems}, vol.~33, pp. 12\,449--12\,460, 2020.

\bibitem{lee2021textless}
A.~Lee, H.~Gong \emph{et~al.}, ``Textless speech-to-speech translation on real data,'' \emph{arXiv preprint arXiv:2112.08352}, 2021.

\bibitem{boito2024mhubert}
M.~{Zanon Boito}, V.~Iyer \emph{et~al.}, ``mhubert-147: A compact multilingual hubert model,'' in \emph{Interspeech 2024}, 2024, pp. 3939--3943.

\bibitem{pratap2024scaling}
V.~Pratap, A.~Tjandra \emph{et~al.}, ``Scaling speech technology to 1,000+ languages,'' \emph{Journal of Machine Learning Research}, vol.~25, no.~97, pp. 1--52, 2024.

\bibitem{chen-etal-2024-towards-robust}
W.~Chen, W.~Zhang \emph{et~al.}, ``Towards robust speech representation learning for thousands of languages,'' in \emph{Proceedings of the 2024 Conference on Empirical Methods in Natural Language Processing}.\hskip 1em plus 0.5em minus 0.4em\relax Miami, Florida, USA: Association for Computational Linguistics, Nov. 2024, pp. 10\,205--10\,224.

\bibitem{Shi2023mlsuperb}
J.~Shi, D.~Berrebbi \emph{et~al.}, ``Ml-superb: Multilingual speech universal performance benchmark,'' in \emph{Interspeech}, 2023.

\bibitem{conneau2023fleurs}
A.~Conneau, M.~Ma \emph{et~al.}, ``Fleurs: Few-shot learning evaluation of universal representations of speech,'' in \emph{2022 IEEE Spoken Language Technology Workshop (SLT)}.\hskip 1em plus 0.5em minus 0.4em\relax IEEE, 2023, pp. 798--805.

\bibitem{meyer2022bibletts}
J.~Meyer, D.~Adelani \emph{et~al.}, ``Bibletts: a large, high-fidelity, multilingual, and uniquely african speech corpus,'' in \emph{Interspeech}.\hskip 1em plus 0.5em minus 0.4em\relax {ISCA}, 2022.

\bibitem{kimanuka2024speech}
U.~Kimanuka, C.~wa~Maina \emph{et~al.}, ``Speech recognition datasets for low-resource congolese languages,'' \emph{Data in Brief}, vol.~52, p. 109796, 2024.

\bibitem{kallaama2024dataset}
E.~Gauthier, A.~Ndiaye \emph{et~al.}, ``Kallaama: A transcribed speech dataset about agriculture in the three most widely spoken languages in senegal,'' in \emph{Proceedings of the Fifth workshop on Resources for African Indigenous Languages (RAIL 2024)}, 2024.

\bibitem{ardila-etal-2020-common}
R.~Ardila, M.~Branson \emph{et~al.}, ``\BIBforeignlanguage{English}{Common voice: A massively-multilingual speech corpus},'' in \emph{\BIBforeignlanguage{English}{Proceedings of the Twelfth Language Resources and Evaluation Conference}}.\hskip 1em plus 0.5em minus 0.4em\relax Marseille, France: European Language Resources Association, May 2020, pp. 4218--4222.

\bibitem{emezue2025naijavoicesdatasetcultivatinglargescale}
\BIBentryALTinterwordspacing
C.~Emezue, T.~N. Community \emph{et~al.}, ``The naijavoices dataset: Cultivating large-scale, high-quality, culturally-rich speech data for african languages,'' 2025. [Online]. Available: \url{https://arxiv.org/abs/2505.20564}
\BIBentrySTDinterwordspacing

\bibitem{barnard2014nchlt}
E.~Barnard, M.~H. Davel \emph{et~al.}, ``The nchlt speech corpus of the south african languages,'' in \emph{Workshop Spoken Language Technologies for Under-resourced Languages (SLTU)}, 2014.

\bibitem{badenhorst2022nchlt}
J.~Badenhorst and F.~De~Wet, ``Nchlt auxiliary speech data for asr technology development in south africa,'' \emph{Data in Brief}, vol.~41, p. 107860, 2022.

\bibitem{doumbouya2021usingradio}
M.~Doumbouya, L.~Einstein \emph{et~al.}, ``Using radio archives for low-resource speech recognition: towards an intelligent virtual assistant for illiterate users,'' in \emph{Proceedings of the AAAI Conference on Artificial Intelligence}, vol.~35, 2021.

\bibitem{valk2021voxlingua107}
J.~Valk and T.~Alum{\"a}e, ``Voxlingua107: a dataset for spoken language recognition,'' in \emph{2021 IEEE Spoken Language Technology Workshop (SLT)}.\hskip 1em plus 0.5em minus 0.4em\relax IEEE, 2021, pp. 652--658.

\bibitem{sikasote23_interspeech}
C.~Sikasote, K.~Siaminwe \emph{et~al.}, ``{Zambezi Voice: A multilingual speech corpus for zambian languages},'' in \emph{Proc. INTERSPEECH 2023}, 2023, pp. 3984--3988.

\bibitem{tang2020multilingual}
Y.~Tang, C.~Tran \emph{et~al.}, ``Multilingual translation with extensible multilingual pretraining and finetuning,'' \emph{arXiv preprint arXiv:2008.00401}, 2020.

\bibitem{alabi-etal-2022-adapting}
J.~O. Alabi, D.~I. Adelani \emph{et~al.}, ``Adapting pre-trained language models to {A}frican languages via multilingual adaptive fine-tuning,'' in \emph{Proceedings of the 29th International Conference on Computational Linguistics}.\hskip 1em plus 0.5em minus 0.4em\relax Gyeongju, Republic of Korea: International Committee on Computational Linguistics, Oct. 2022, pp. 4336--4349.

\bibitem{xu2024seamless}
J.~Xu, M.~Wu \emph{et~al.}, ``Seamless language expansion: Enhancing multilingual mastery in self-supervised models,'' \emph{arXiv preprint arXiv:2406.14092}, 2024.

\bibitem{douze2024faisslibrary}
M.~Douze, A.~Guzhva \emph{et~al.}, ``The faiss library,'' 2024.

\bibitem{ott2019fairseq}
M.~Ott, S.~Edunov \emph{et~al.}, ``fairseq: A fast, extensible toolkit for sequence modeling,'' in \emph{Proceedings of NAACL-HLT 2019: Demonstrations}, 2019.

\bibitem{caubriere2024africacentric}
A.~Caubri{\`e}re and E.~Gauthier, ``Africa-centric self-supervised pretraining for multilingual speech representation in a sub-saharan context,'' in \emph{5th Workshop on African Natural Language Processing}, 2024.

\bibitem{kudo-richardson-2018-sentencepiece}
T.~Kudo and J.~Richardson, ``{S}entence{P}iece: A simple and language independent subword tokenizer and detokenizer for neural text processing,'' in \emph{Proceedings of the 2018 Conference on Empirical Methods in Natural Language Processing: System Demonstrations}.\hskip 1em plus 0.5em minus 0.4em\relax Brussels, Belgium: Association for Computational Linguistics, Nov. 2018, pp. 66--71.

\bibitem{speechbrain}
M.~Ravanelli, T.~Parcollet \emph{et~al.}, ``{SpeechBrain}: A general-purpose speech toolkit,'' 2021, arXiv:2106.04624.

\bibitem{goyal-etal-2022-flores}
N.~Goyal, C.~Gao \emph{et~al.}, ``The {F}lores-101 evaluation benchmark for low-resource and multilingual machine translation,'' \emph{Transactions of the Association for Computational Linguistics}, vol.~10, pp. 522--538, 2022.

\bibitem{ma24c_interspeech}
M.~Ma, Y.~Koizumi \emph{et~al.}, ``Fleurs-r: A restored multilingual speech corpus for generation tasks,'' in \emph{Interspeech 2024}, 2024, pp. 1835--1839.

\bibitem{ahia-etal-2024-voices}
O.~Ahia, A.~Aremu \emph{et~al.}, ``Voices unheard: {NLP} resources and models for {Y}or{\`u}b{\'a} regional dialects,'' in \emph{Proceedings of the 2024 Conference on Empirical Methods in Natural Language Processing}.\hskip 1em plus 0.5em minus 0.4em\relax Miami, Florida, USA: Association for Computational Linguistics, Nov. 2024, pp. 4392--4409.

\end{thebibliography}
